\documentclass[lettersize,journal]{IEEEtran}
\usepackage{amsmath,amsfonts}
\usepackage{algorithmic}
\usepackage{algorithm}
\usepackage{array}
\usepackage[caption=false,font=normalsize,labelfont=sf,textfont=sf]{subfig}
\usepackage{textcomp}
\usepackage{stfloats}
\usepackage{url}
\usepackage{verbatim}
\usepackage{graphicx}
\usepackage{cite}
\usepackage{graphicx} 
\usepackage{gensymb}
\usepackage{xcolor}
\begin{document}

\title{A peristaltic soft, wearable robot for compression and massage therapy}

\author{Mengjia Zhu$^{1}$, Adrian Ferstera$^{1,2}$, Stejara Dinulescu$^{1}$, Nikolas Kastor$^{1}$, Max Linnander$^{1}$, \\  Elliot W. Hawkes$^{1}$, and Yon Visell$^{1}$
\thanks{$^{1}$Media Arts and Technology Program, Department of Electrical and Computer Engineering, Department of Mechanical Engineering, California NanoSystems Institute, University of California, Santa Barbara, USA
        {\tt\small yonvisell@ucsb.edu}}%
\thanks{$^{2}$Mechanical Engineering Program, Department of Mechanical Engineering, ETH Zurich, Switzerland}%
\thanks{Manuscript received XXX, 2022; revised XXX, 2022.}}

\markboth{IEEE/ASME TRANSACTIONS ON MECHATRONICS, 2022}%
{Shell \MakeLowercase{\textit{et al.}}: A Sample Article Using IEEEtran.cls for IEEE Journals}


\maketitle

\begin{abstract}
Soft robotics is attractive for wearable applications that require conformal interactions with the human body. Soft wearable robotic garments hold promise for supplying dynamic compression or massage therapies, such as are applied for disorders affecting lymphatic and blood circulation. 
In this paper, we present a wearable robot capable of supplying dynamic compression and massage therapy via peristaltic motion of finger-sized soft, fluidic actuators. We show that this peristaltic wearable robot can supply dynamic compression pressures exceeding 22 kPa at frequencies of 14 Hz or more, meeting requirements for compression and massage therapy. A large variety of software-programmable compression wave patterns can be generated by varying frequency, amplitude, phase delay, and duration parameters. 
We first demonstrate the utility of this peristaltic wearable robot for compression therapy, showing fluid transport in a laboratory model of the upper limb. We theoretically and empirically identify driving regimes that optimize fluid transport.
We second demonstrate the utility of this garment for dynamic massage therapy.
These findings show the potential of such a wearable robot for the treatment of several health disorders associated with lymphatic and blood circulation, such as lymphedema and blood clots. 
\end{abstract}


\section{Introduction}

Emerging soft robotic technologies are becoming widely investigated for applications in health and medicine.    
Soft, wearable robotic systems are well-suited for delivering dynamic compression or massage therapy.  Such therapies, when administered by a health professional, are effective in treating  blood and lymphatic circulatory disorders, including chronic venous insufficiency (CVI), venous ulcers, lymphedema, and blood clots \cite{duffield2016effects,partsch2012compression}, as well as for treating musculoskeletal injuries \cite{crane2012massage}.

Compression therapy devices are expected to meet functional requirements arising from therapeutic standards that help ensure efficacy and safety or otherwise conform to norms of massage practice. Such guidelines are necessarily application-specific. Compression therapies requiring gross occlusion of venous blood flow in the lower limbs demand substantial compression pressures, amounting to 3.3 kPa, 8.0 kPa, or 9.3 kPa in the supine, sitting, or standing positions respectively  \cite{partsch2005calf}. Intermittent compression therapies, such as are used for lymphedema, require pressure variations to be supplied at frequencies of at least 1 Hz \cite{payne2018force, delis2000optimum}. In other applications, massage therapies are supplied for mild pain relief, easing of musculoskeletal tension, or touch elicited (i.e., haptic) emotional comfort. Examples include sports, Swedish, kinesiologic, or deep tissue massage.  Such practices can involve the application of light to firm pressures, ranging from less than 12 kPa to 30 kPa or more \cite{roberts2011effects}.  

Wearability introduces further requirements that arise from ergonomics and safety.  To ensure comfort, compression garments should be soft, low-profile, and compliant.  To avoid pain and avoid injury or lesions damaging soft tissues, static and dynamic forces should be distributed to avoid stress concentrations. For devices that emulate manual massage, it may be appropriate that compression is supplied via individually actuatable modules the size of a finger or hand. 

Based on these requirements for compression therapies, the two existing predominant tools are graduated compression stockings (GCS) and intermittent pneumatic pressure (IPC) devices. 
GCS promote venous flow by applying a constant compression pressure to the limb to reduce venous caliber, preventing the static accumulation of blood \cite{partsch2012compression}. GCS are convenient and low-cost while lacking the capability of dynamic pressure adjustment. 
IPC devices consist of single or multiple pneumatic chambers that are inflated to exert compression pressure on the limb \cite{chen2001intermittent}. Compared with GCS, IPC treatment features dynamic pressure modes with a wide range of pressure delivery (1-20 kPa) \cite{moran2015systematic}. Studies have shown that the use of IPC instead of GCS is associated with lower incidents of venous thromboembolism (VTE) \cite{arabi2013use}. Researchers have also shown that sequential actuation modes and high pressure in a multi-chamber device generate a higher lymphatic flow than other modes of treatment \cite{kitayama2017real}. 

IPC devices rely on the inflation of the bladders to exert compression pressure on the skin, making it difficult to control the contact area for efficient mechanical coupling with blood vessels. 
It is also challenging for a user to move around when wearing them because the bladders used in IPC devices are usually bulky.
The use of smaller bladders may reduce the clinical efficacy of the device due to the smaller volume of blood being expelled \cite{comerota2011intermittent}.

\begin{figure*}[ht]
  \centering
  \includegraphics[width=\textwidth]{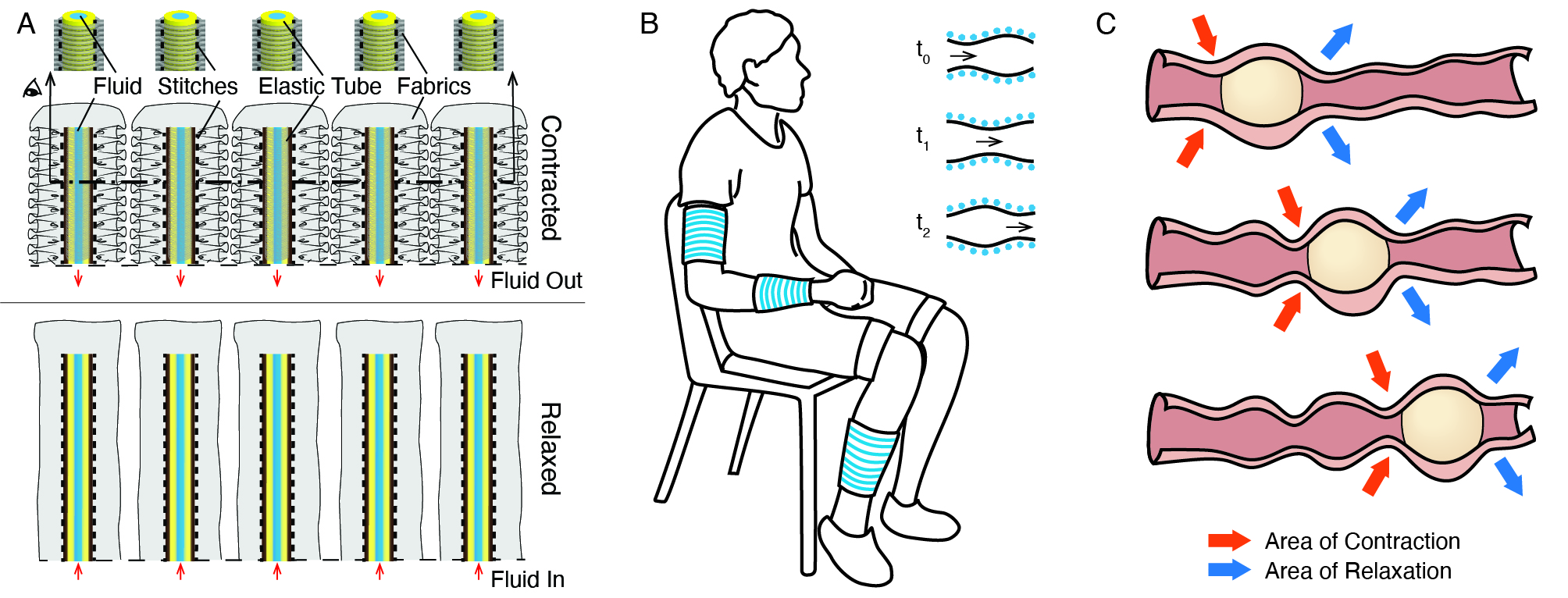}
  \caption{A. FFMS actuators are used for the peristaltic wearable robot design. These actuators are arranged in parallel to mimic peristalsis. B. The peristaltic wearable robot can be used in various locations on human limbs, including the upper arm, forearm, and lower limb. The insert demonstrates the propagation of the peristaltic wave generated. C. Peristaltic transport in biological organs such as the digestive tract involves sequential contraction and relaxation of muscles.}
  \label{fig:concept}
\end{figure*}
\setlength{\textfloatsep}{15pt}

Prior researchers have investigated the modeling, control, and device designs for therapeutic massaging to overcome these limitations in IPC devices.
Guan et al. proposed an evidence-based fluid-structure coupling model for the interaction between IPC soft actuator and lower limb, providing guidelines for improving the efficacy of compression therapy for venous flow promotion \cite{guan2020fluid}.
To address control challenges for pneumatic operations, 
Payne et al. incorporated closed-loop force control in pneumatic actuators with similar sizes to IPC modules \cite{payne2018force}.
Zhao et al. proposed a programmable and self-adaptive dynamic pressure delivery system with a matrix of soft sensors for IPC therapies \cite{zhao2020programmable}.

In addition to IPC devices, several soft, wearable robotic systems have been investigated for supplying forces to the body \cite{zhu2022review, thalman2020review} via pneumatic muscles \cite{wu2019wearable}, shape memory alloys \cite{papadopoulou2019affective, otsuka1999shape}, or motor-driven tendons \cite{kang2019exo, hofmann2018design}.
For example, Suarez et al. created a system for robotic lymphatic drainage using three pneumatic bending fabric actuators to compress and laterally push fluid in a flexible tube \cite{suarez2018soft}.
Yoo et al. developed Z-folded pneumatic
actuator modules to generate normal and shear forces for lymphedema massaging \cite{yoo2019wearable}. 

Pneumatic actuation has been widely used in such applications, but is often energetically inefficient, due to thermodynamic losses arising from gas compression cycles \cite{zhu2020fluidic}. Pneumatic actuators also store substantial energy in gas compression, which can pose hazards upon soft material failure, due to rapid energy release. Shape memory alloy actuators perform mechanical work via a thermally actuated phase change; consequently, such actuators are energetically inefficient and operate at low speeds constrained by heat transport \cite{otsuka1999shape}. 
Other designs utilize tendon- or Bowden-cable transmissions. Such wearable devices require careful design to avoid painful or hazardous stress concentration \cite{kang2019exo}. 

To overcome some of these limitations, we present a wearable system based on fluidic fabric muscle sheets (FFMS), described in recent publications by the authors \cite{zhu2020fluidic, zhu2020pneusleeve, zhu2022review}. FFMS supply forces via soft, hydraulic transmissions integrated into finger-sized compact textile matrices (Fig. \ref{fig:concept} A). They achieve greater energy efficiency and safety than many pneumatic designs, avoid stress concentration through the use of soft materials, and are capable of supplying large forces (scalable based on application requirements) and of rapid actuation at speeds exceeding 10 Hz, meeting requirements for dynamic compression and massage therapies in many application areas.

Leveraging the merits of FFMS actuators, we demonstrate a soft wearable robot that generates massage-like motions on the human limb through dynamic compression waves based on peristalsis (Fig. \ref{fig:concept} B).
Peristalsis is a transport mechanism produced via sequential squeezing of muscles, as in the esophagus \cite{paterson2006esophageal}, uterus \cite{kuijsters2017uterine}, ureter \cite{edmond1970human}, or blood vessels \cite{misra2002peristaltic}. Media within a peristaltic channel is transported via wave motion of the channel walls (Fig. \ref{fig:concept} C). 
Peristaltic transport has been previously applied in robotic systems for liquid pumping \cite{jaffrin1971peristaltic} or swallowing robots  \cite{chen2013soft}. When applied to the exterior of the body, peristalsis can supply forces similar to those used in compression and massage therapies \cite{roberts2011effects}.

We present the design, fabrication, and modeling of the peristaltic wearable robot and the associated control system. 
We characterize the compression pressure, dynamic frequency responses, and motion patterns generated by the peristaltic wearable robot.
Results show that the system we designed can supply compression pressure as large as 22 kPa with frequencies up to 14 Hz, meeting the requirements for 
compression and massage therapy
\cite{partsch2005calf, payne2018force, delis2000optimum}.
Our proof-of-concept demonstration for compression therapy shows that this peristaltic wearable robot is capable of driving venous flow in an artificial limb with a flow rate up to 1 mL/min. The flow rate increases linearly with peristaltic wavelength and frequency, in agreement with theoretical models we derive based on prior literature.
Finally, we demonstrate 
that our wearable robot can deliver
a wide range of massage-like motions on the forearm. This work provides a controllable and efficient method for supplying compression massage that could treat
venous and lymphatic inefficiencies, as well as a means for therapeutic massage that could provide comfort or reduce anxiety.

\section{Device Design and Operating Principle}

\subsection{Wearable Design and Fabrication}

The peristaltic compression wearable aims to create dynamic compression force patterns on the wearer’s limbs (Fig. \ref{fig:concept} B). 
It consists of eight fluidic fabric muscle sheets (FFMS) actuators and leverages several benefits of this soft robotic technology \cite{zhu2020fluidic}.
FFMS actuators are intrinsically compliant and can be scaled to sizes as small as human fingers.
The modular arrangement of these actuators offers extra flexibility for personalized configurations. In addition, these actuators can be controlled efficiently with hydraulics \cite{zhu2020fluidic} and are
capable of generating a large variety of compression patterns
that are easy to be detected and distinguished by humans \cite{zhu2020pneusleeve}.
These characteristics make FFMS an ideal candidate that meets the design requirements for compression therapy.

The operating principle of FFMS resembles Inverse Pneumatic Artificial Muscles (IPAM) \cite{hawkes2016design}. When fluid pressure is high, the actuator extends to its longest state. When the fluid pressure decreases, the actuator contracts and exerts forces on the payload (Fig. \ref{fig:concept} A). When wrapped around a human limb circumferentially, each actuator is capable of supplying localized compressive pressure to the limb. This working principle is intrinsically different from the IPC devices, such that a compact-sized FFMS actuator may still supply large compression pressures.
The parallel arrangement of multiple actuators makes it possible to provide spatial-temporal compression patterns, including wave-like patterns that produce peristalsis.

We fabricated the actuators using the methods described in Zhu et al. \cite{zhu2020fluidic}. The geometry of the soft latex tubing used inside the fabric conduits affects the fluid pressure required to fully operate the actuator, as well as the maximum compression force generated \cite{zhu2020fluidic}. 
The actuator length should be designed such that the actuator can wrap around the human limb and produce compression pressure of at least 9.3 kPa for compression therapy.
Based on these requirements and the analytical modeling in Zhu et al. \cite{zhu2020fluidic}, we specified the tubing to have an outer diameter of 4.8 mm, and an inner diameter of 3.2 mm. We selected the stitching length as 177 mm and the wrinkling ratio was 2.5.
The fabricated actuators were then pre-pressurized and hand-sewn onto a stretchable sleeve for comfort using a zig-zag pattern that preserves stretchability.

\subsection{Design and Modeling of the Fluid Driving System}
    
\begin{figure}[t]
  \centering
  \includegraphics[width=\columnwidth]{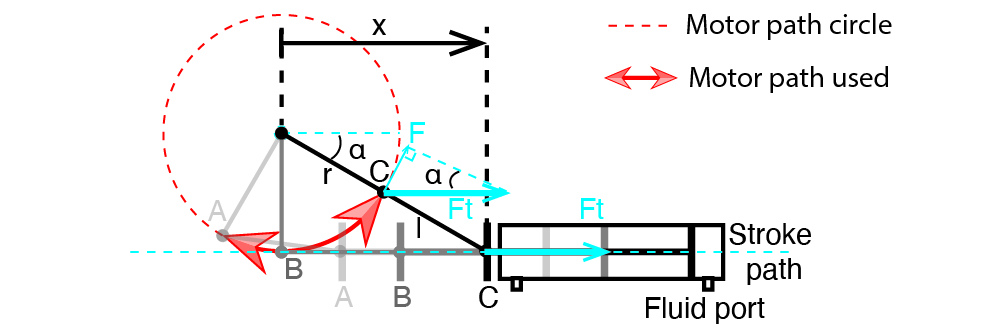}
  \caption{Schematic design of the slider-crank driving mechanism.}
  \label{fig:slidercrank}
\end{figure}

\begin{figure*}[ht]
  \begin{center}
  \includegraphics[width=0.95\textwidth]{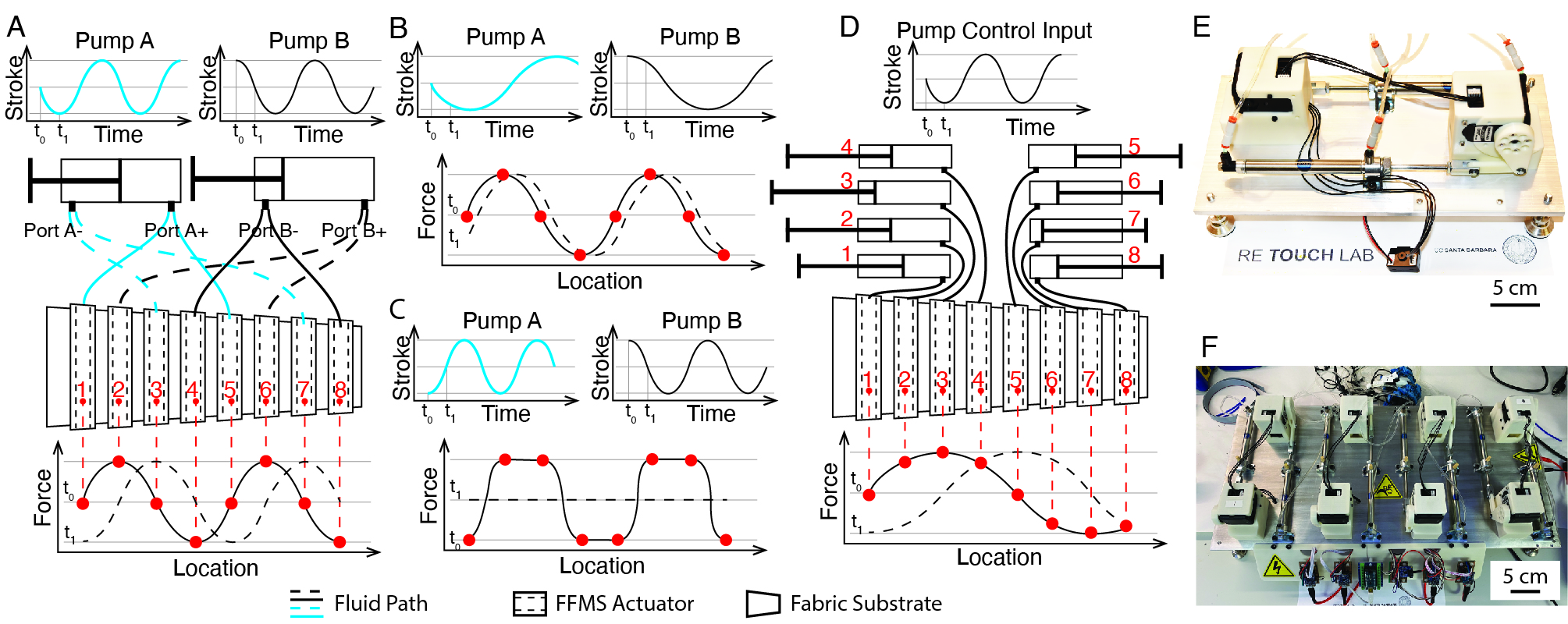}
  \caption{Various control input designs can be used to generate spatial-temporal compression wave patterns. A. Two pistons can be used to control eight actuators for maximum efficiency, with spatial compression forces coupled at locations routed to the same motor. The sinusoidal displacements with a phase delay of $\pi/2$ on the top denote an example of the input signal for both pistons. The plot on the bottom illustrates the resulting spatial profile of the compression force at times $t_0$ and $t_1$. 
  The two-piston driving system can also compose other spatial-temporal compression force patterns such as B and C, with the top figures showing the piston input signals, and the bottom figures showing the composed spatial waves.
  D. The driving system can be expanded to eight pistons to control each actuator independently. Arbitrary spatial compression force waves can be generated. E. The prototype of the two-piston driving system. F. The prototype of the eight-piston driving system.}
  \label{fig:actuation}
  \end{center}
\end{figure*}

The system uses compact, servo-driven hydraulic cylinders to drive the fluid (water).
The power requirements for the motors are determined by the hydraulic pressure and the speed of the actuation. A slider-crank linkage converts the rotary motion to linear motion (Fig. \ref{fig:slidercrank}). The minimum stroke position occurs when the crank $r$ is co-linear with the connecting rod $l$ (position C). It is the initial position of the motor when fluid pressure is maximum. This design uses $l = r$ such that the minimum operation angle $\alpha$ between the crank and the horizontal line is $ 30^{\circ}$. The range of $\alpha$ is adjustable to accommodate a different fluid volume. For our actuator, the maximum $\alpha$ needed is $120^{\circ}$ (position A).
When the crank is vertical, the connecting rod is co-linear with the stroke path (position B). 
With this design, the torque $\tau$ of the motor can be related to the force acting on the piston rod of the hydraulic cylinder $F_t$ by:

\begin{equation}\label{crankTorque}
    \tau = r F_t \sin\alpha, \ 30^{\circ} \leq \alpha \leq 120^{\circ}\
\end{equation}

In a quasi-static condition, $F_t = PA$, where $P$ is the quasi-static fluid pressure and $A$ is the inner cross-sectional area of the hydraulic cylinder.
The relationship between the crank angle $\alpha$ and the piston location $x$ is obtained from the kinematics of a fourbar offset slider-crank \cite{norton2004}:

\begin{equation}\label{pistonPosition}
    x = r\cos{\alpha} + l\sqrt{1-\left(\frac{r\cdot\sin{\alpha}-r}{l}\right)^2}, \\
    \ 30^{\circ} \leq \alpha \leq 120^{\circ}\
\end{equation}

Our two-motor design is sufficient to produce peristaltic motions via coupled hydraulic channels (Fig. \ref{fig:actuation} A, E). 
As the piston of the hydraulic cylinder moves back and forth, the hydraulic pressure increases in one port but decreases in the other port. With two motors, the eight actuators can be connected to the four ports, with two channels sharing one port. When two sinusoidal signals with a phase delay of $\pi/2$ are fed into the two motors, a compression force pattern similar to a sinusoidal wave can be generated spatially along the wearable robot (Fig. \ref{fig:actuation} A). Different input signals result in distinct spatial-temporal compression force patterns (Fig. \ref{fig:actuation} B, C).

\begin{figure}[ht]
  \centering
  \includegraphics[width=0.48\textwidth]{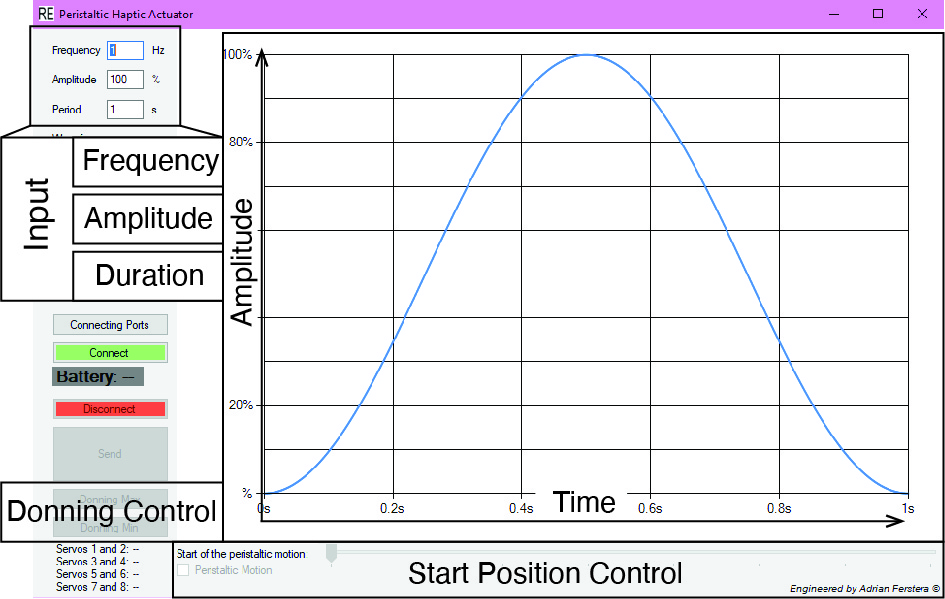}
  \caption{The graphic user interface generates a user-defined waveform with input parameters such as frequency and amplitude. The interface also includes a two-step donning design of the device.}
  \label{fig:motordesign}
\end{figure}

\begin{figure}[t]
  \centering
  \includegraphics[width=\columnwidth]{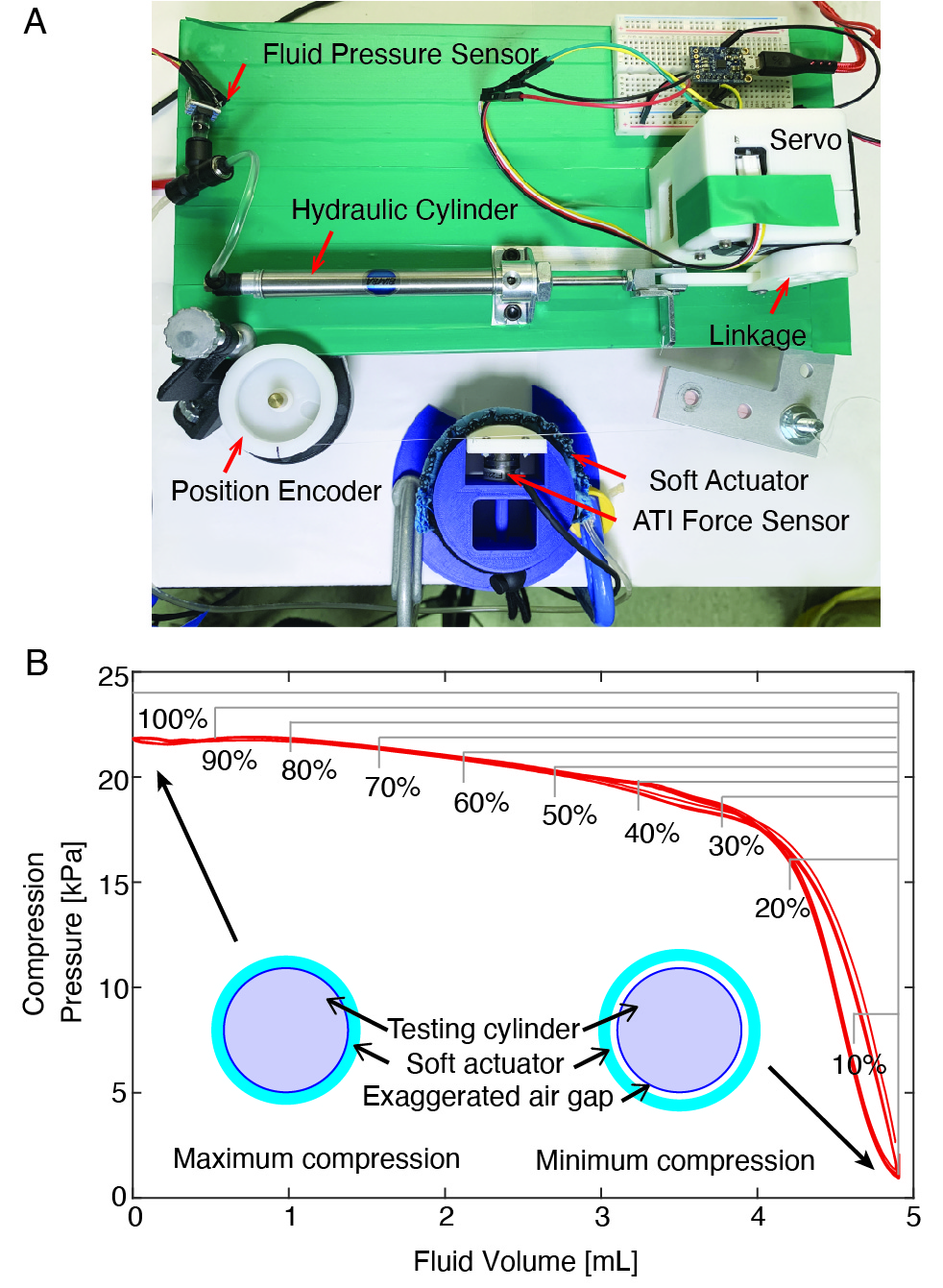}
  \caption{Evaluation of the force generated by a single actuator. A. Compression force testing set-up for a single actuator. B. Relationship between fluid volume and compression pressure follows a similar curve for different operation strokes ranging from 10\% to 100\% with the margins marked in the plot. Note that the minimum compression pressure occurs when the fluid volume is maximum. As the fluid is withdrawn from the actuator, the compression pressure increases. The inserted diagrams illustrate the extreme conditions for maximum and minimum compression pressure to occur.}
  \label{fig:forcetesting}
\end{figure}

The two-motor compact design uses a minimal number of motors at the cost of coupled compression forces. In many other situations, each hydraulic channel needs to be controlled independently.
We designed a second driving system using eight pistons to allow independent control over individual channels to generate arbitrary compression force patterns (Fig. \ref{fig:actuation} D, F).

\subsection{Software Development}
\label{(GUI) Design}

Each piston is driven by a programmable high torque servo (HerklueX DRS 0602, DST Robot) that is controlled by computer via a synchronous serial communication.
A software graphical user interface (GUI) enables the design of compression patterns via frequency, amplitude, duration, and phase parameters (Fig. \ref{fig:motordesign}). 
The motors can be powered using a rechargeable Lipo battery (14.8 V, 9 A).
The software also implements a two-step donning process. First, pistons are positioned to fully pressurize one chamber in the hydraulic cylinder, allowing actuators connected to that chamber to be fastened to the arm of the user.
Then, pistons are positioned to fully pressurize the other chamber, allowing the rest of the actuators to be fastened.

\section{Mechanical Characterization of the Dynamic Compression System}

Various peristaltic, all-in-phase, and sequential squeezing spatial-temporal compression patterns can be delivered through time-varying signals for a single actuator, as well as multiple actuators at different locations.
In this section, we evaluate the compression force generated by a single actuator, the dynamic frequency response of the system, and the large range of dynamic compression motion patterns that the peristaltic wearable robot is capable of generating.

\subsection{Single Actuator Compression Force Testing}
We characterized the compression force of a single actuator by wrapping it around a cylindrical testing fixture with a force sensor (ATI Nano17; ATI Industrial Automation) (Fig. \ref{fig:forcetesting} A). 
The force sensor measures part of the compression force exerted on the arc segment with a central angle of 68.9 degrees. 
The diameter of the cylinder in the testing setup is 63.7 mm to mimic the size of a human limb. 
Different design parameters of the testing fixture, such as the cylinder radius, can affect the compression pressure measurement. 
Applying the hoop stress equation yields the following expression for the compression pressure:
\begin{equation}\label{eq:compressionPressure}
P_c = \frac{t_{act}\sigma_{act}}{R} = \frac{F}{A_c},
\end{equation}
where $P_c$ represents the compression pressure, $t_{act}$ and $\sigma_{act}$ are the thickness and the axial stress of the actuator, $R$ is the radius of the cylinder wrapped by the actuator, $F$ is the force measured from the ATI nano 17, $A_c$ is the contact area between the actuator and the testing cylinder. A position encoder measured the instantaneous stroke of the hydraulic cylinder and a pressure sensor 
was used to monitor hydraulic pressure. All sensing signals were recorded synchronously using a real-time computer-in-the-loop system with a sampling rate of 100 Hz (QPIDe; Quanser, Inc., with Simulink, The MathWorks, Inc.).

Results showed that compression pressure decreased with increasing fluid volume in the actuator, which is consistent with the model presented in Zhu et al. \cite{zhu2020fluidic} (Fig. \ref{fig:forcetesting} B).
The resting state of the actuator is when the fluid volume is maximum, leading to minimum compression pressure. When fluid volume is withdrawn from the actuator, the compression pressure increases.
This relationship was approximately linear both before and after fluid volume of 4 mL, with a larger slope for fluid volume between 4 mL and 5 mL. The maximum compression pressure decreased with operation stroke range, as the margins illustrated in Fig. \ref{fig:forcetesting} B.

\begin{figure*}
  \centering
  \includegraphics[width=\textwidth]{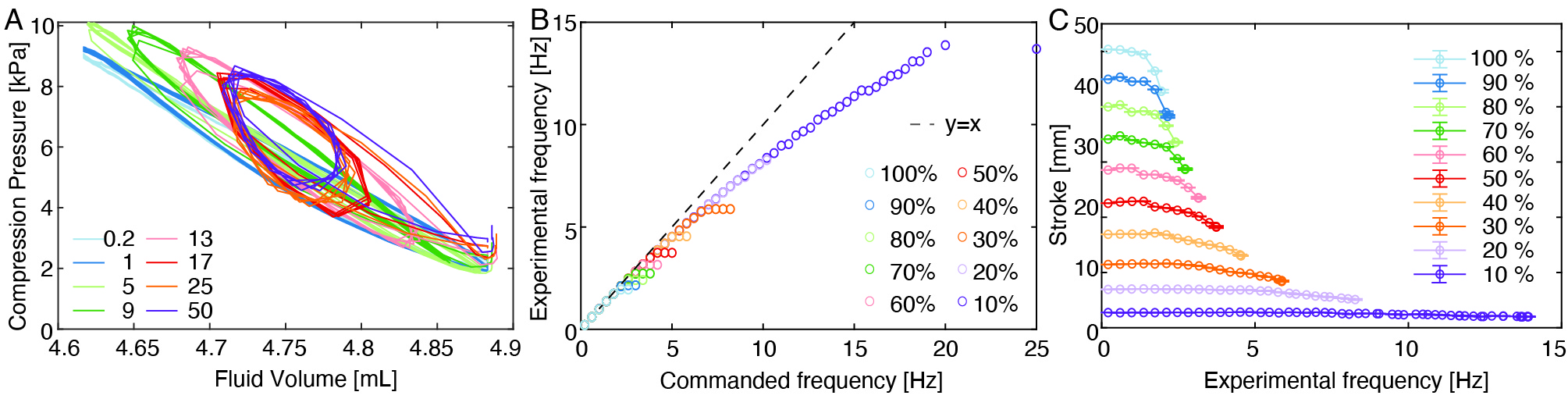}
  \caption{Frequency response of the system. A. The hysteresis loops presented in the relationship between fluid volume and compression pressure show energy loss during the actuation. The energy loss increases with increasing frequency. Legend: commanded frequency in Hz.
  B.  Legend: different operation stroke range in \%. The experimental frequency reaches a plateau for all stroke ranges. The smaller the stroke range, the higher the experimental frequency achieved. Overall, the system is able to operate at frequencies up to about 20 Hz despite some discrepancies between commanded frequency and experimental frequency.
  C. As the stroke magnitude gets higher, the drop-off in stroke occurs at lower frequencies. The magnitude remains almost flat across all frequencies for 10\% operation stroke range.}
  \label{fig:freqplot}
\end{figure*}

\begin{figure*}
  \centering
  \includegraphics[width=\textwidth]{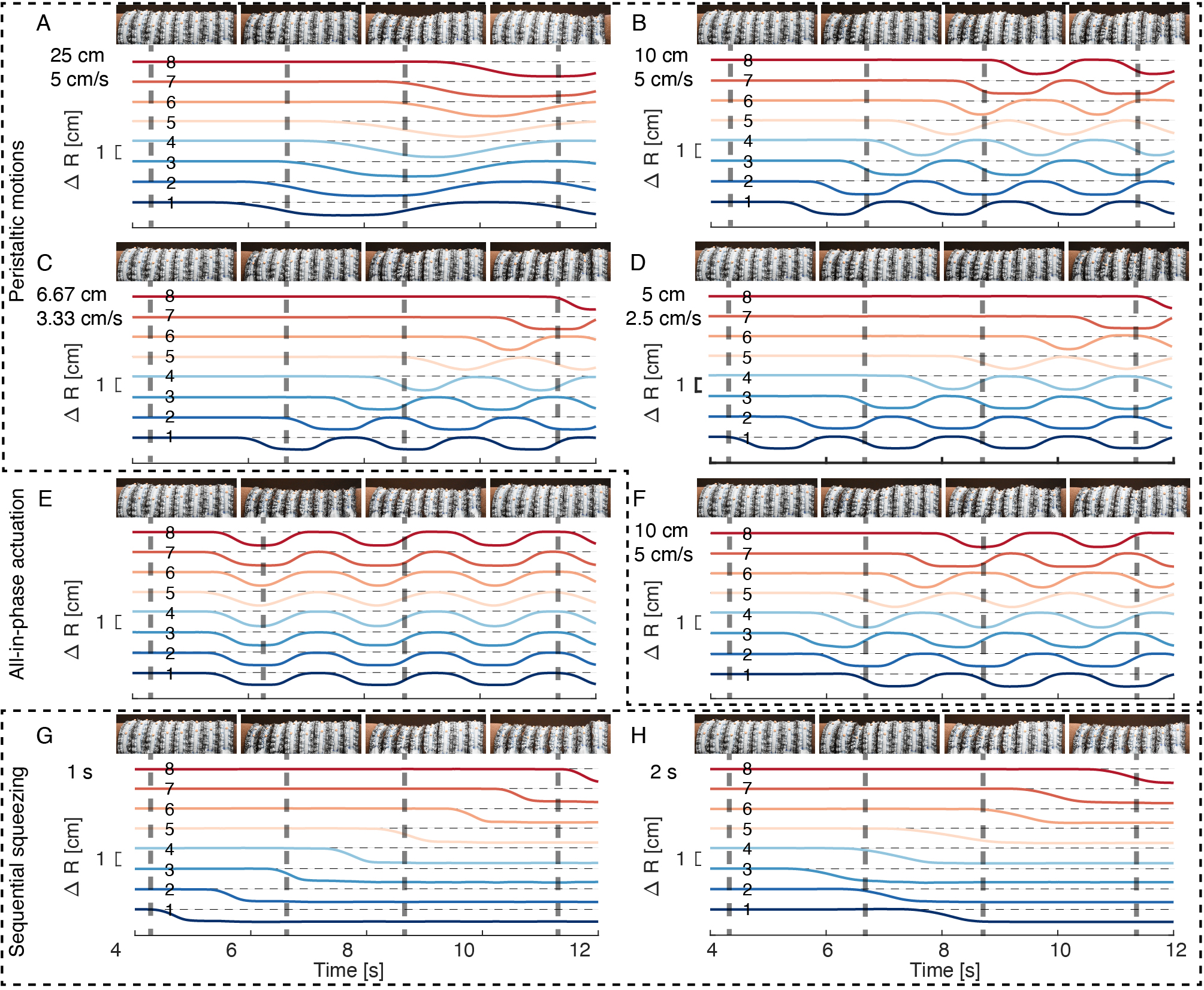}
  \caption{Motion capture results for various dynamic compression patterns including peristaltic motions (A-D, F), all-in-phase actuation (E), and sequential squeezing (G-H). The texts at the top left corner of each subfigure denote the wavelength and wave speed of each peristaltic pattern. For the sequential squeezing patterns, the texts show the squeezing time of each actuator. 
  }
  \label{fig:motiondata}
\end{figure*}

\begin{figure*}
  \centering
  \includegraphics[width=\textwidth]{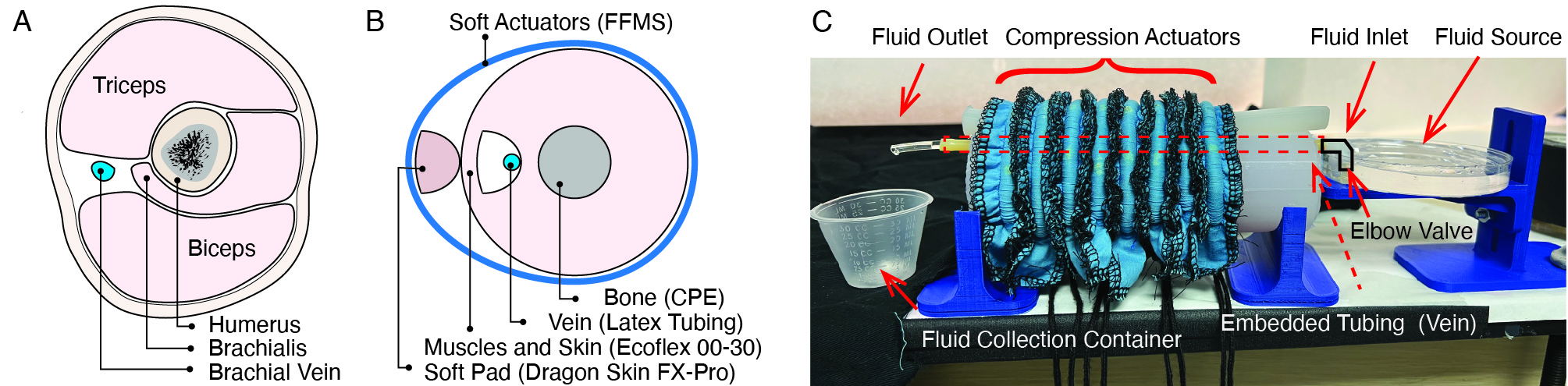}
  \caption{The wearable robot is configured to promote hemodynamic circulation in a limb model via peristaltic transport. A. A simplified illustration of the cross-section through the middle of upper arm \cite{eycleshymer1911cross}. B. The cross-section of the arm model used for the experiment. C. Testing apparatus for the peristaltic transport. }
  \label{fig:PeristalticTransport}
\end{figure*}

\subsection{Dynamic Frequency Response of the System}

We evaluated the system's dynamic frequency response at frequencies ranging from 0.2 Hz to 50 Hz with stroke amplitudes from 10\% to 100\%.
The stroke of the piston was measured, and the actual frequency (experimental frequency) was obtained using spectrum analysis of the data. Fig. \ref{fig:freqplot} A shows the results for the stroke amplitude of 10\%. Compression pressure decreased as frequency increased up to 17 Hz, above which the compression pressure and the fluid volume had similar ranges. 
Results also revealed that the system was capable of operating at frequencies up to 20 Hz (Fig. \ref{fig:freqplot} B).
Motor power constrained the stroke amplitude in that the lower the maximum stroke used, the higher frequency achieved. 
The magnitude decrement was more prominent for larger maximum stroke (Fig. \ref{fig:freqplot} C), while little magnitude decrement was observed when operating with 10\% stroke.

\subsection{Dynamic Compression Motion Patterns}

We characterized the dynamic spatial compression motion patterns by measuring shape deformation with  an optical motion capture system (Flex 13; OptiTrack).
Twenty-four reflective markers (5 mm diameter) were distributed on the surface of the wearable robot, with three on each actuator. Compression patterns were generated and the dynamic positions of each marker are measured. Fig. \ref{fig:motiondata} shows the average radius changes of the three markers on each actuator over time. Five peristaltic motions were composed by varying wavelength, wave speed, and the starting actuator (Fig. \ref{fig:motiondata} A-D, F), along with an all-in-phase synchronized compression pattern (Fig. \ref{fig:motiondata} E). Two sequential squeezing patterns were also evaluated with squeezing times of 1 s and 2 s respectively (Fig. \ref{fig:motiondata} G-H). These results mechanically validated that the wearable robot was capable of generating different compression patterns on a soft body.

\section{Applications for Compression Therapy and Therapeutic Massage}
We demonstrated applications of the peristaltic wearable robot for compression therapy to promote blood circulation and for therapeutic massage therapy.

\subsection{Compression Therapy for Blood Flow Promotion}

Peristaltic action on a flexible tube can introduce fluid flow, thus, can be used to promote blood flow in the veins, to treat various medical conditions \cite{duffield2016effects,partsch2012compression,crane2012massage}. 
To demonstrate the ability of the peristaltic compression robot to promote blood circulation, we designed and fabricated an upper arm model (Fig. \ref{fig:PeristalticTransport} B) that mimics the anatomy of the human upper arm (Fig. \ref{fig:PeristalticTransport} A) \cite{eycleshymer1911cross}. The muscles (triceps, biceps,  and brachialis muscles) and skin were realized using a soft silicone material (Ecoflex 00-30, shore hardness 00-30, Smooth-On, Inc., Macungie, PA). A rigid 3D printed bone structure was fabricated in CPE (Ultimaker filament). A latex tube was placed adjacent to the bone,  representing the brachial vein. A soft strip made of stiffer silicone with a shore hardness of 2A (Dragon Skin FX-Pro, Smooth-On, Inc., Macungie, PA) was used to direct compression pressure to the vein.

\begin{figure*}
  \centering
  \includegraphics[width=\textwidth]{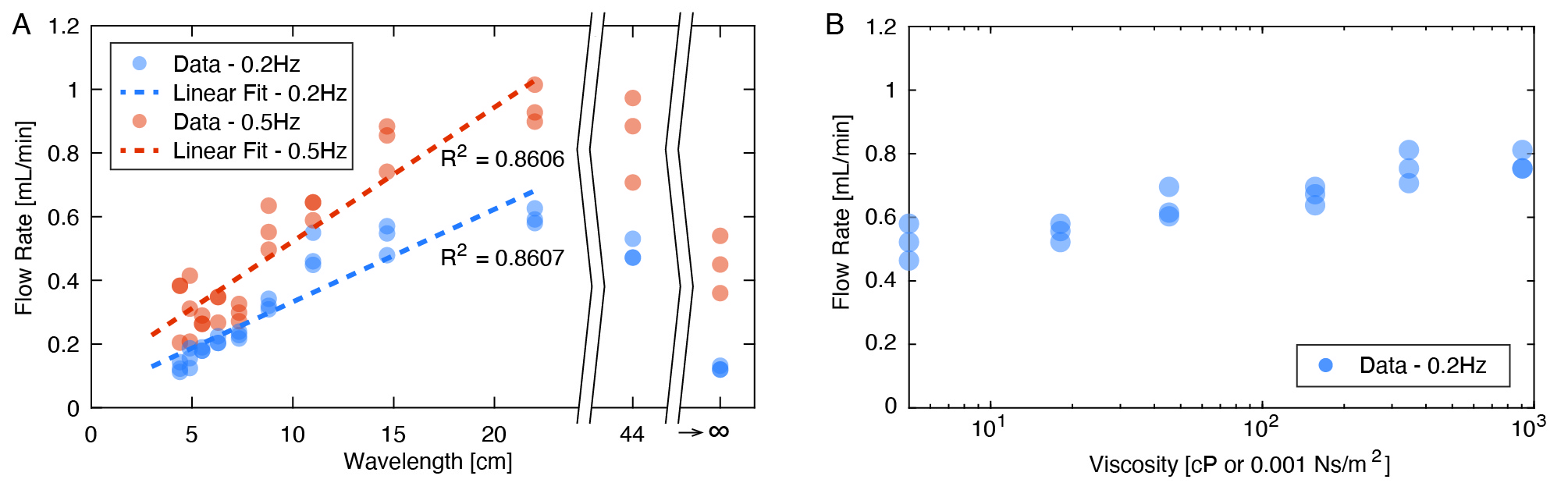}
  \caption{A. Experimental results of the flow rate showed a linear relationship with the wavelength, in agreement with the theoretical prediction (equation \ref{eq:flowRatef}). B. Results showed that the viscosity had a weak effect on the flow rate, agreeing with the analytical predictions in \cite{barton1968peristaltic,jaffrin1973inertia}.  }
  \label{fig:PeristalticTransport_Data}
\end{figure*}

The limb apparatus is shown in Fig. \ref{fig:PeristalticTransport} C. 
Vegetable glycerin served as the fluid for transport in the artificial vein, as it can be mixed with water for different viscosity values \cite{cheng2008formula}. 
Peristaltic compression patterns were generated when each motor rotated sinusoidally with an onset delay time of $\Delta t$ between adjacent actuators:

\begin{equation}\label{eq:inputsignal}
\alpha=A\sin(\,2\pi f (t+  (n-1)\Delta t)\,),
\end{equation}
where $\alpha$ is the rotation degree for the motor, $n$ indexes the actuator channel, $t$ is the time variable, $A$ and $f$ are the amplitude and frequency of the sinusoidal wave respectively. The spatial wave generated by all the actuators is described as:
\begin{equation}\label{eq:spatialWave}
y(x,t) = A'\cos(\frac{2\pi}{\lambda}x - 2\pi f (t-\Delta t)),
\end{equation}
where $y$ and $x$ are the vertical and horizontal position variables,  $A'$ and $\lambda$ are the amplitude and wavelength of the spatial wave respectively.

The peristaltic compression robot was wrapped around the limb model, with a spacing of about 1.1 cm between adjacent actuators. During peristaltic actuation, a flow was induced, transporting fluid from the source through an elbow connector to the collection container (Fig. \ref{fig:PeristalticTransport} C). 
From the theory of peristaltic flow in flexible tubes \cite{barton1968peristaltic}, when the pressure change over one wavelength $\Delta P_\lambda = 0$, 
the time average peristaltic flow $\bar{Q}$ in a tube depends on the average radius of the tube $a$, wave amplitude $b$, and wavespeed $c$ as follows:
\begin{equation}\label{eq:flowRatec}
 \bar{Q}=\frac{\pi c b^2}{2}
 \frac{16a^2-b^2}{2a^2+3b^2}
\end{equation}
When the frequency $f$ of the peristaltic wave is constant,  the time average flow increases with wavelength $\lambda$ due to the increase in unoccluded volume:
\begin{equation}\label{eq:flowRatef}
 \bar{Q}=\frac{\pi \lambda f b^2}{2}  \frac{16a^2-b^2}{2a^2+3b^2}
\end{equation}
Frequencies of 0.2 Hz and 0.5 Hz were tested.
Different peristaltic compression patterns were generated with onset delay time stepping between 0 ms (all-in-phase actuation) and 1125 ms (90\degree  ~phase shift) by 125 ms (9\degree  ~phase shift). Each wave pattern was repeated three times, and the fluid volume collected in the container was measured.

\begin{table}
    \centering
    \caption{Theoretical viscosity and Reynolds number of glycerin-water mixtures with different glycerin concentrations in mass $C_m$.}
    \begin{tabular}{cccc p{1cm} p{1cm} p{2.8cm} p{2cm}} 
    \hline 
        $C_m$
        & Density
        & Dynamic viscosity [cP]
        & Reynolds number
        \\ [0.1cm] \hline
        0.5 
        & 1.12
        & 5.57
        & 0.6427
        \\
         0.7
        & 1.17
        & 20.88
        & 0.1799
        \\ 
         0.8
        & 1.20
        & 53.96
        & 0.0714
        \\ 
        0.9
        & 1.23
        & 193.24
        & 0.0204
        \\
        0.95
        & 1.24
        & 438.10
        & 0.0091
        \\ 
        1
        & 1.26
        & 1,178.64
        & 0.0034
        \\ \hline
         \end{tabular}\\
    \label{tab:viscosity}
\end{table}

We found that the flow rate was higher for peristaltic motions than all-in-phase actuation when the wavelength was larger than about 5 cm. The time-average flow rate increased with the wavelength following a linear relationship (Fig. \ref{fig:PeristalticTransport_Data} A), in agreement with theoretical predictions (equation \ref{eq:flowRatef}). The maximum flow rate was achieved when the wavelength was about 22 cm. When the wavelength further increased to 44 cm, the flow rate decreased. This may be due to the boundary conditions in the testing configuration, as the total length of actuation (about 9.2 cm) was much smaller than the wavelength of 44 cm. When the wavelength was very large, the wave pattern resembled all-in-phase actuation and the fluid reflux counteracting with the peristaltic flow was large. This decrement for wavelengths above 22 cm was also presented when frequency increased to 0.5 Hz. The overall flow rate for 0.5 Hz was slightly larger than 0.2 Hz, in agreement with theoretical predictions (equation \ref{eq:flowRatef}).

The dynamic viscosity of glycerin at temperature 22\degree  ~is about 1179 cP, which is much larger than the normal blood viscosity values (between 3.3 cP and 5.5 cP) \cite{nader2019blood}. Blood is also a non-Newtonian fluid whose viscosity changes depending on the hemodynamic conditions. Here, we evaluated the effect of fluid viscosity on induced flow rate. 
The viscosity of the transported fluid was varied by mixing glycerin with water at different weight percentages. The theoretical viscosity of the glycerin-water mixture was obtained using published results \cite{cheng2008formula} (Table \ref{tab:viscosity}).

Experimental results showed that viscosity had a weak effect on the flow rate (Fig. \ref{fig:PeristalticTransport_Data} B). The flow rate decreased as the viscosity decreased. This finding is consistent with predictions in the literature that show that viscosity have little effect on the flow rate when viscous forces are dominant (small Reynolds numbers) \cite{barton1968peristaltic,jaffrin1973inertia}.

\subsection{Therapeutic Massage with a Haptic Sleeve}
Wave-like peristaltic compression can also be used to emulate manual massage therapy for stress reduction and muscle relaxation. Papadopoulou et al. developed an affective sleeve that produces sequential compression and warmth on the forearm using shape memory alloys (SMAs) \cite{papadopoulou2019affective}.

\begin{figure}
  \centering
  \includegraphics[width=0.48\textwidth]{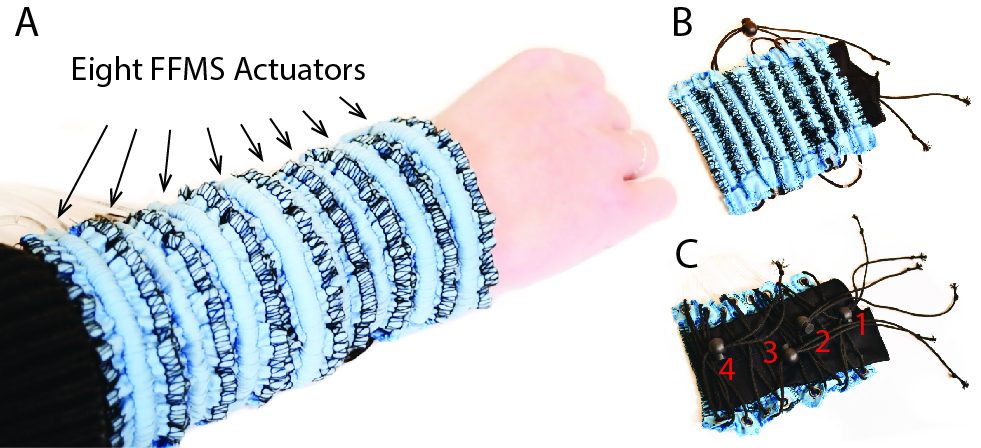}
  \caption{A. Our peristaltic wearable robot on user's forearm for therapeutic massage. B. The actuator side of the wearable robot. C. The closure side of the wearable robot.}
  \label{fig:devicepicture}
\end{figure}

We adapted our peristaltic wearable robot to supply massaging patterns (Fig. \ref{fig:devicepicture}).
Different massage patterns may be generated by varying actuation frequency, amplitude (operation stroke range), the phase time delay between adjacent actuators, and the initialization location for the wave.
A supplementary video shows several examples illustrating patterns of haptic massage feedback generated by the device. 

\section{Discussion and Future Work}
This article presents a peristaltic soft, wearable robot for robotic compression therapy. Although IPC has remained the main tool for automated compression therapy for decades, little device renovation has been done to improve the spatial arrangement due to the limitations of pneumatic operation. We present a novel peristaltic wearable robot that has finely distributed modules with a width comparable to the size of a finger for localized compression. We also implement a modular hydraulic actuation system for efficient control compared to commonly used pneumatic wearables. Each module is able to generate compression pressure up to 22 kPa and frequency up to 14 Hz, meeting the requirements for compression therapy and 
therapeutic haptic massage
\cite{partsch2005calf, payne2018force, delis2000optimum}. With programmable frequency, amplitude, phase delay, and duration, it is capable of rendering a large range of spatial-temporal wave patterns.  We further demonstrate the application of the peristaltic wearable robot for blood flow promotion and therapeutic haptic massage.

The theoretical and experimental results demonstrate the device's ability to drive fluid flow in a model limb via peristaltic transport, and provide a quantitative model that parameters governing the peristaltic pattern to the flow rate and fluid properties.  These findings indicate that such systems may hold substantial promise for the treatment of lymphatic and blood circulation inefficiencies, which are associated with several health disorders. 
The model arm used in our experiments is in reality quite different than a real arm.
Future translational research aimed at designing and evaluating the clinical efficacy of such a peristaltic robotic sleeve is warranted.

\section*{Acknowledgments}
The authors acknowledge support from the U.S. National Science Foundation Awards 1751348 and 2002529 to Visell, 2139319 to Dinulescu, and 1944816 to Hawkes. The authors acknowledge the use of the Microfluidics Laboratory and Innovation Workshop within the California NanoSystems Institute, supported by the University of California, Santa Barbara and the University of California, Office of the President.

\bibliographystyle{IEEEtran} 
\bibliography{IEEEabrv,DynamicCompression}

\newpage

\section*{Biography Section}
 
\vspace{11pt}
\vspace{-33pt}
\begin{IEEEbiography}[{\includegraphics[width=1in,height=1.25in,clip,keepaspectratio]{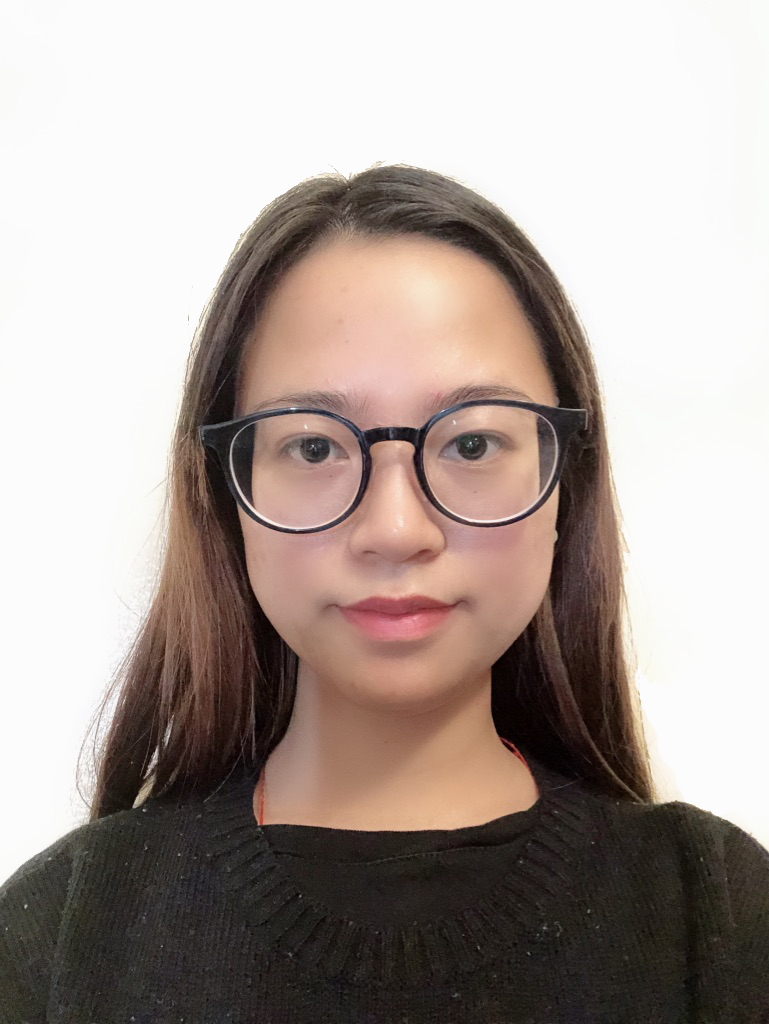}}]{Mengjia Zhu}
received the B.Eng. (Hons.) degree in apparel design and engineering from Soochow University, Suzhou, China, in 2015, and the M.S. degree in materials science and engineering from Arizona State University, Tempe, AZ, USA, in 2017. 

She is currently working toward the Ph.D. degree in media arts and technology at the University of California, Santa Barbara, Santa Barbara, CA, USA, under the supervision of Prof. Yon Visell in the RE Touch Lab. Her research interests lie in the area of functional wearables including wearable haptics, soft robotics, and assistive technologies.
\end{IEEEbiography}

\vspace{11pt}
\vspace{-33pt}
\begin{IEEEbiography}[{\includegraphics[width=1in,height=1.25in,clip,keepaspectratio]{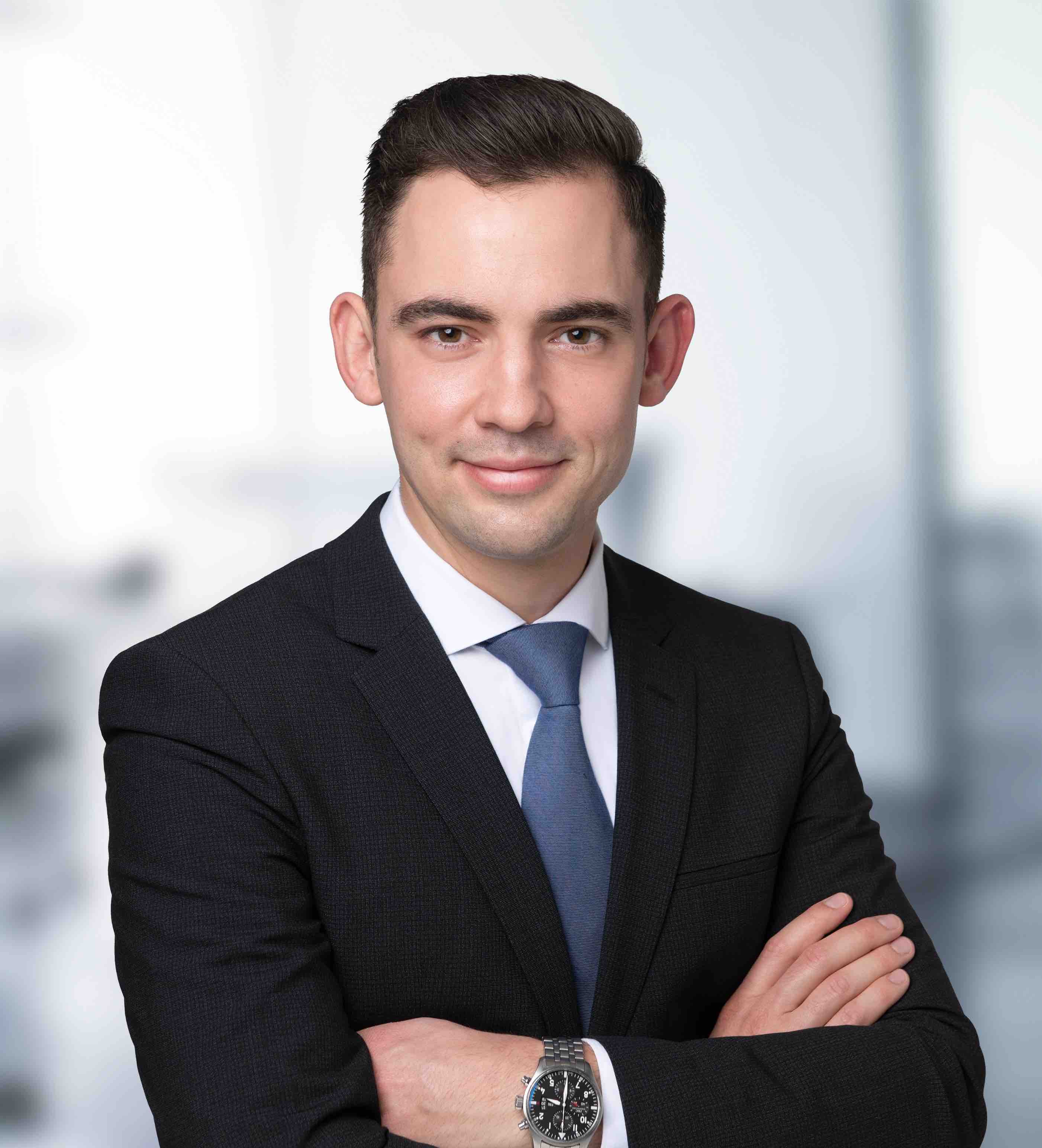}}]{Adrian Ferstera}
received his M.Sc in Mech Eng with a major in Robotics, Systems and Control from ETH Zurich. During his time at ETH Zurich, he focused on software development for any robotic application under the supervision of Prof. Bradley Nelson in the Multi-Scale Robotics Laboratory. He is currently working for a software consultancy company which focuses on software development for novel banking applications.
\end{IEEEbiography}

\vspace{11pt}
\vspace{-33pt}
\begin{IEEEbiography}[{\includegraphics[width=1in,height=1.25in,clip,keepaspectratio]{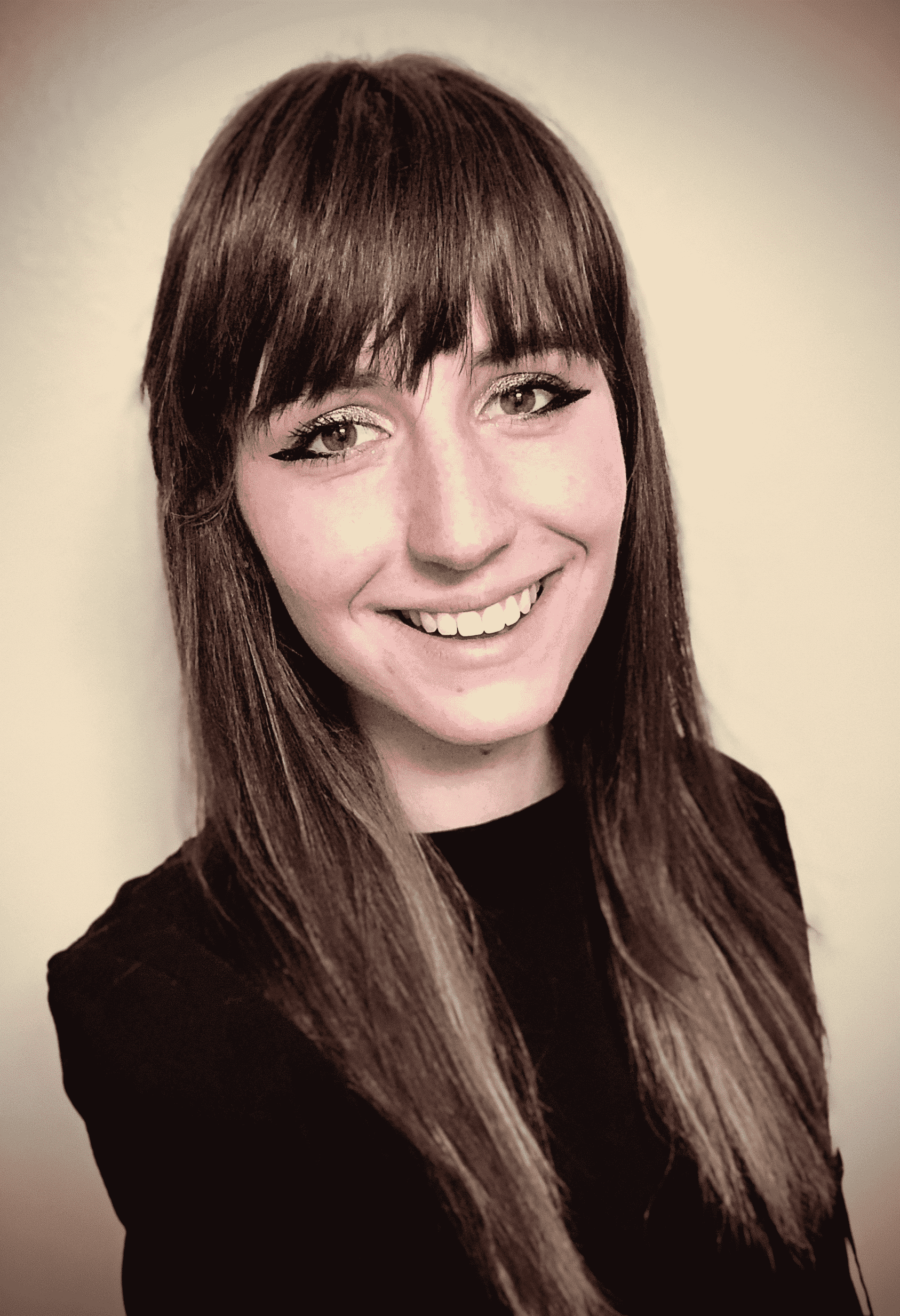}}]{Stejara Iulia Dinulescu}
received the B.S. degree in Psychology (with Distinction), the B.A. degree in Creative Computation, and the B.A. degree in Studio Arts from Southern Methodist University, Dallas, TX, USA, in 2019, with minors in Neuroscience and Cognitive Science. She is currently a Ph.D. candidate in the Media Arts and Technology department at the University of California, Santa Barbara, Santa Barbara, CA, USA, working in the RE Touch Lab and mentored by Prof. Y. Visell. Her research interests include studying the perceptual and mechanical properties of interpersonal touch and developing wearable haptic and soft robotic devices for reproducing touch experiences remotely.
\end{IEEEbiography}

\vspace{11pt}
\vspace{-33pt}
\begin{IEEEbiography}[{\includegraphics[width=1in,height=1.25in,clip,keepaspectratio]{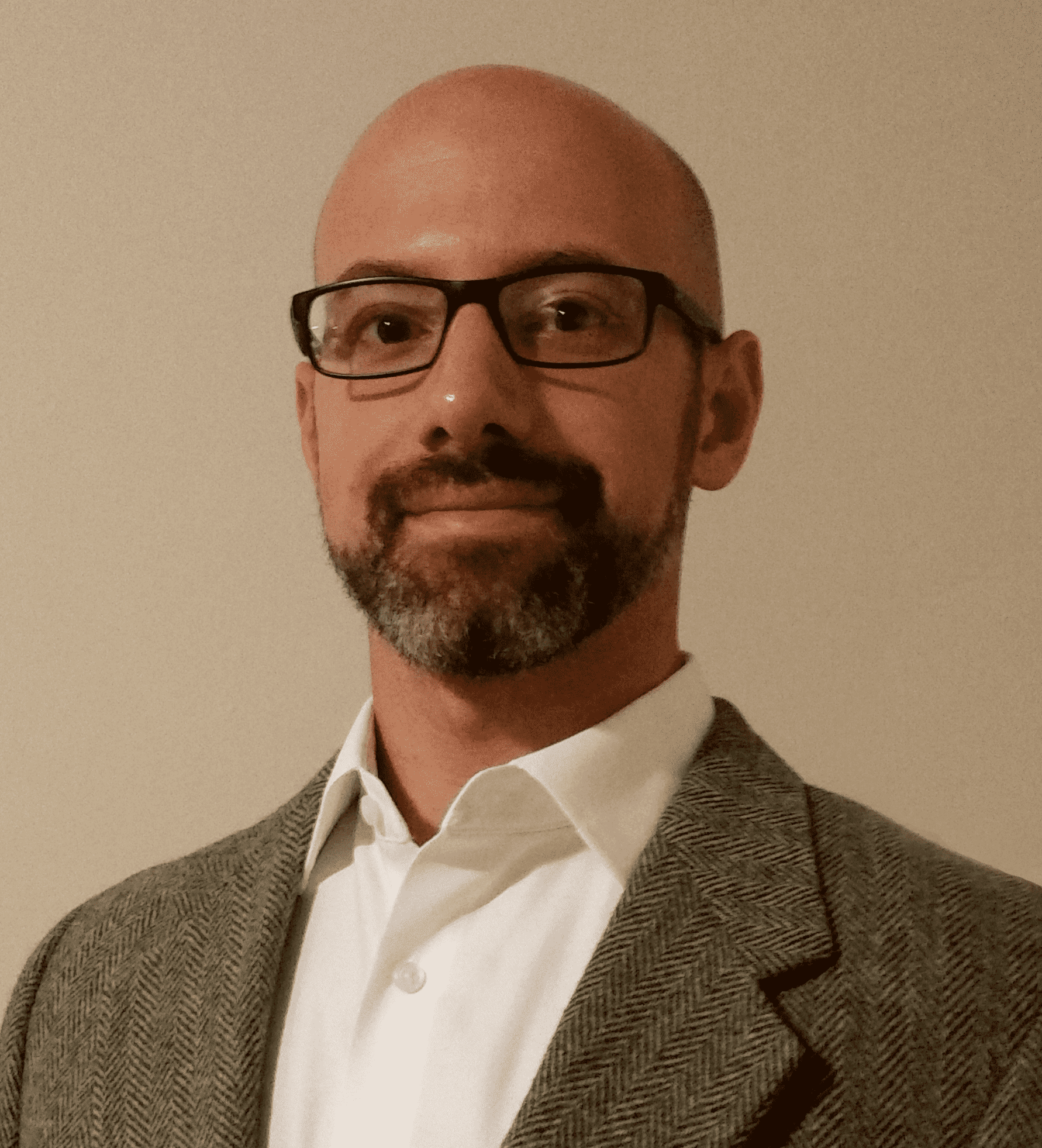}}]{Nikolas Kastor}
holds a Ph.D. in Mechanical Engineering from Tufts University and is a Fellow of the Tufts University Soft Material Robotics National Science Foundation (NSF) Integrative Graduate Education and Research Traineeship (IGERT).  Dr. Kastor contributes over 13 years of research, design, manufacturing, and project management experience in the fields of soft material robotics, haptics, MEMS sensors, commercial transducers, and consumer products; where he has published numerous papers and patents.  He has conducted business and presented research in North America, Europe, and Asia.  Dr. Kastor has participated in the NSF Innovation Corps (I-Corps) customer discovery training for a wearable robotics application and is currently self-employed. During this project, Nikolas was a postdoctoral scholar at the University of California Santa Barbara working on soft haptic, sensing, and actuation systems with the RE Touch Lab.
\end{IEEEbiography}

\vspace{11pt}
\vspace{-33pt}
\begin{IEEEbiography}[{\includegraphics[width=1in,height=1.25in,clip,keepaspectratio]{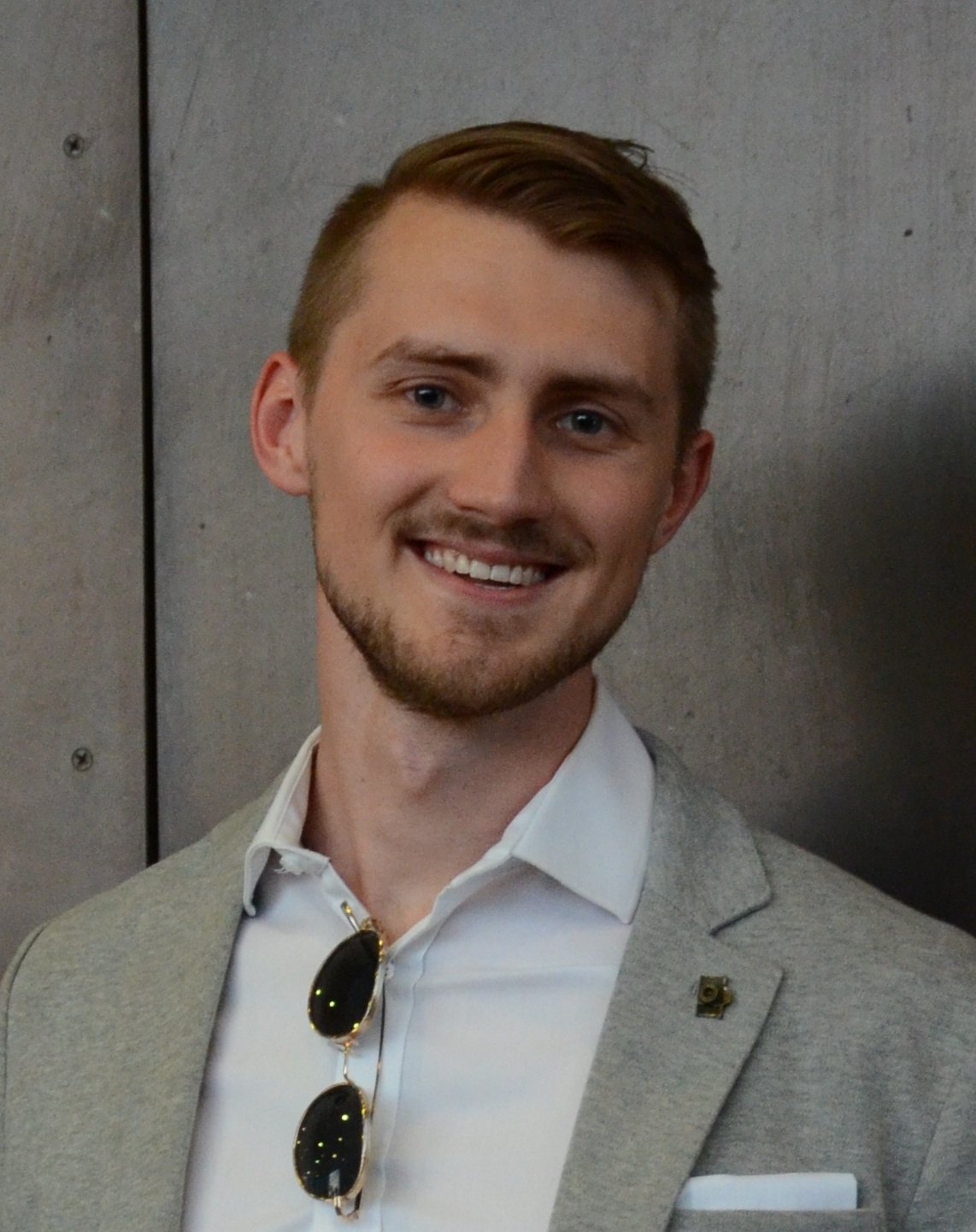}}]{Max Linnander} received the B.Sc. degree in physics from Arizona State University, Tempe, AZ, USA, in 2021. He is currently working toward the Ph.D. degree in mechanical engineering with the University of California, Santa Barbara, Santa Barbara, CA, USA, under the supervision of Prof. Y. Visell.
His research interests include optics, acoustics, and haptics.

\end{IEEEbiography}

\vspace{11pt}
\vspace{-33pt}
\begin{IEEEbiography}[{\includegraphics[width=1in,height=1.25in,clip,keepaspectratio]{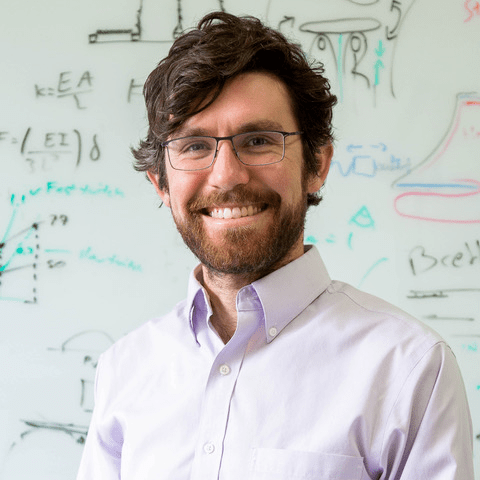}}]{Elliot Wright Hawkes} (Member, IEEE) received the A.B. degree with highest honors in mechanical engineering from Harvard University, Cambridge, MA, USA; the M.S. degree in mechanical engineering from Stanford University, Stanford, CA, USA; and the Ph.D. degree in mechanical engineering from Stanford University under the supervision of Prof. M. Cutkosky, in 2009, 2012, and 2015, respectively.

He is an Assistant Professor with the Department of Mechanical Engineering, University of California, Santa Barbara, USA. His research interests include compliant robot design, mechanism design, nontraditional materials, artificial muscles, directional adhesion, and growing robots.

Dr. Hawkes recently received the National Science Foundation CAREER Award and the NASA Early Faculty Career Award.
\end{IEEEbiography}

\vspace{11pt}
\vspace{-33pt}
\begin{IEEEbiography}[{\includegraphics[width=1in,height=1.25in,clip,keepaspectratio]{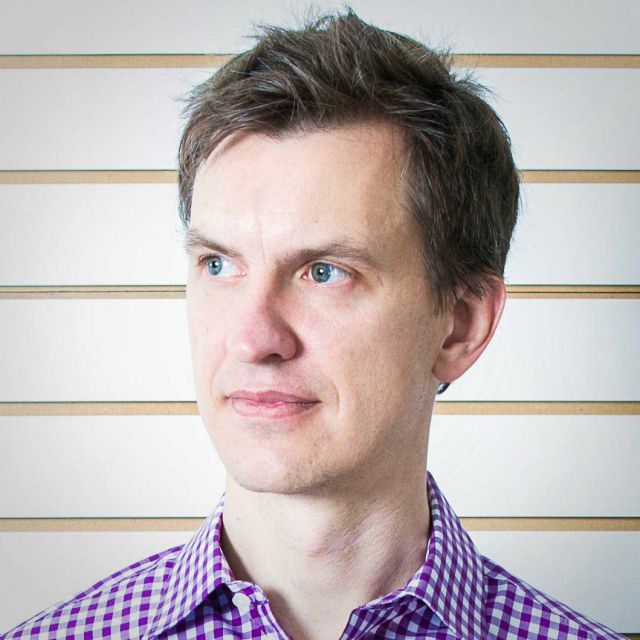}}]{Yon Visell}
received the B.A. degree in physics from
the Wesleyan University, Middletown, CT, USA, in
1995, and the M.A. degree in physics from The
University of Texas, Austin, Austin, TX, USA, in
1999, and the Ph.D. degree in electrical and computer engineering from McGill University, Montreal, QC, Canada, in 2011.

He is currently an Associate Professor with the
University of California, Santa Barbara, Santa Barbara, CA, USA, in the Media Arts and Technology
Program, Department of Electrical and Computer
Engineering, and Department of Mechanical Engineering (by courtesy). His academic interests include haptics, soft robotics, and soft electronics. He was a Postdoctoral Scholar with Sorbonne University, Paris. Prior to his Ph.D. studies, Visell spent several years in industry positions, including that of digital signal processing (DSP) developer for Ableton Live. He has authored and coauthored more than 75 scientific works.

Dr. Visell was the recipient of several awards for work presented at prominent academic conferences. He received a Google Faculty Research Award in 2016, a Hellman Family Foundation Faculty Fellowship in 2017, and a U.S. National
Science Foundation CAREER award in 2018. He is an Associate Editor for the IEEE ROBOTICS AND AUTOMATION LETTERS and serves as General Co-Chair of the 2022 IEEE Haptics Symposium.
\end{IEEEbiography}

\vfill

\end{document}